\documentclass[11pt]{article}
\usepackage{acl}

\usepackage{times}
\usepackage{latexsym}

\usepackage[T1]{fontenc}

\usepackage{listings}
\usepackage{tcolorbox}
\tcbuselibrary{breakable}
\usepackage{xcolor}

\definecolor{myorange}{RGB}{230,120,20}

\usepackage[utf8]{inputenc}

\usepackage{microtype}

\usepackage{inconsolata}

\usepackage{graphicx}
\usepackage{subcaption}
\usepackage{booktabs}      
\usepackage{caption}       
\usepackage[dvipsnames]{xcolor}
\usepackage[table]{xcolor}
\usepackage{amsmath}       
\usepackage{natbib}        

\usepackage{amsmath}
\usepackage{amssymb}

\newcommand{\down}[1]{\textcolor{downred}{\scriptsize$\downarrow$#1}}
\usepackage{arydshln}
\usepackage{fontawesome}
\definecolor{lightpurple}{RGB}{248, 243, 206} 
\definecolor{darkgreen}{rgb}{0.0, 0.5, 0.0}   
\definecolor{darkred}{rgb}{0.8, 0.0, 0.0}     
\usepackage{multirow}
\usepackage{makecell}
\usepackage{enumerate}
\usepackage{enumitem}
\usepackage{pifont}
\usepackage{graphicx}
\usepackage{bbding}
\usepackage[table,dvipsnames]{xcolor}
\usepackage{colortbl}
\usepackage{xspace}

\definecolor{ForestGreen}{RGB}{34,139,34}
\definecolor{myyellow}{RGB}{181, 181, 27}

\definecolor{darksalmon}{rgb}{0.91, 0.59, 0.48}
\definecolor{emerald}{rgb}{0.31, 0.78, 0.47}
\definecolor{greenpigment}{rgb}{0.0, 0.65, 0.31}
\definecolor{amaranth}{rgb}{0.9, 0.17, 0.31}

\definecolor{downred}{HTML}{C0392B} 

\definecolor{iris}{rgb}{0.35, 0.31, 0.81}
\definecolor{uu}{rgb}{0.95, 0.51, 0.51}
\definecolor{spirodiscoball}{rgb}{0.06, 0.75, 0.99}
\definecolor{lightblue}{RGB}{235,245,255}

\title{\textsc{APEx}: Distillation of Agent Procedural Experience for Adaptive Deep Research Question Answering}

\author{
  \textbf{Jie Ding}\textsuperscript{1}\thanks{\,Equal contribution.} \quad
  \textbf{Rui Sun}\textsuperscript{1}\footnotemark[1] \quad
  \textbf{Xinyuan Zhang}\textsuperscript{3} \quad
  \textbf{Zeyu Zhang}\textsuperscript{2} \quad
  \textbf{Xin Liu}\textsuperscript{2}\thanks{\,Corresponding author.}
  \\
  \textsuperscript{1}University of Science and Technology of China \\
  \textsuperscript{2}Chinese Academy of Sciences \quad
  \textsuperscript{3}ByteDance Inc. \\
  \texttt{\{jieding25,rrsun\}@mail.ustc.edu.cn}, \texttt{xliu@gia.cas.cn}
}

\begin{document}
\maketitle
\begin{abstract}
Deep research agents augment large language models with external tools to answer complex, long-horizon questions through multi-turn reasoning. Learning from prior experience is crucial for continual improvement, yet existing methods either retrieve verbose task-specific traces that burden decision-making, 
or distill procedural skills that remain decoupled from downstream policy adaptation. We propose \textsc{APEx}, a hierarchical experience utilization framework that organizes interaction history into instance-level trajectory memories and category-level procedural skills, and couples them through a closed-loop architecture of Executor, Distiller, and Planner. The three modules are optimized via a three-stage alternating GRPO training paradigm, enabling reward-guided skill distillation rather than fixed-prompt generation. At test time, distilled skills serve as procedural priors for online Planner adaptation through skill-guided test-time reinforcement learning, allowing ground-truth-free self-improvement with skill-alignment regularization to prevent policy drift. Experiments on 7 benchmarks demonstrate that \textsc{APEx} achieves state-of-the-art performance, surpassing GPT-5.4 by 14.7 points and the strongest memory-augmented baseline by 3.0 points. Code is available at~\url{https://github.com/J-Ding519/APEx}.
 
\end{abstract}
\section{Introduction}

Deep research agents (DRAs) augment large language models (LLMs) with external tools to tackle complex, long-horizon, and open-ended questions through multi-turn answering~\cite{sun2025simpledeepsearcher,sun2026topodim,li2026openresearcher,zhu2026marco}.
These tasks often involve multi-step tool orchestration~\cite{wu2026webdancer,li2026webthinker}, iterative information retrieval~\cite{jin2025search,wu2025mmsearch}, and multi-source synthesis~\cite{chen2025mindsearch}, making learning from prior experience essential.
Without effective experience utilization, DRAs risk repeating past mistakes instead of building on prior successes~\cite{wang2024agent,wang2025mirix}.
This raises a central question: how can deep research agents continually improve through long-term interaction experience?

\begin{figure}
    \centering
    \includegraphics[width=1\linewidth]{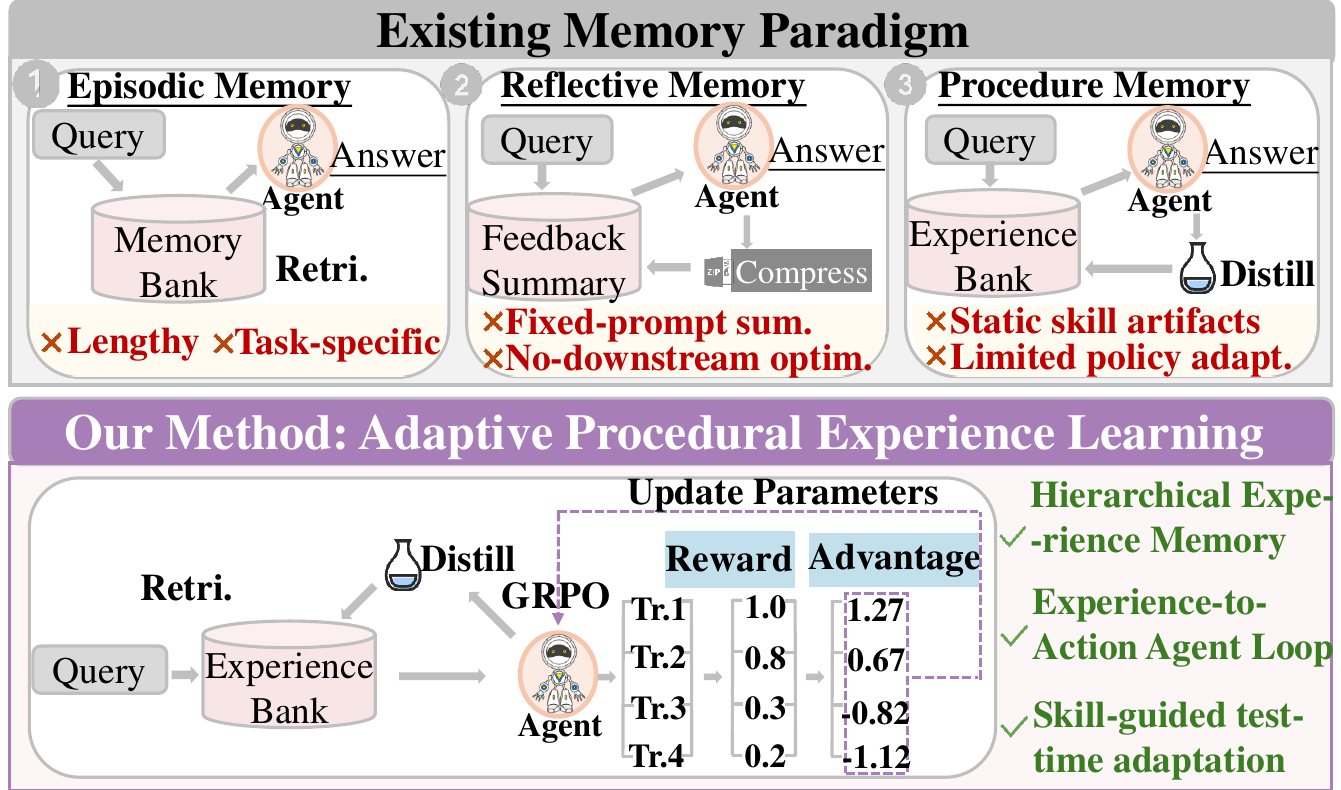}
    \caption{Comparison between existing memory-based agent methods and our adaptive procedural experience learning framework.}
    \label{fig:moti}
\end{figure}

Recent work has attempted to answer this question by equipping agents with memory mechanisms that store, retrieve, and reuse prior experience~\cite{zhou2025memento,qiao2026memory}.
Episodic memory methods store and retrieve past trajectories or task instances~\cite{zhong2024memorybank,zhao2024expel}, but retrieved experiences are often lengthy, task-specific, and noisy, forcing agents to infer transferable strategies at decision time. 
Reflective memory methods compress feedback into textual summaries~\cite{shinn2023reflexion,tan2025prospect}, 
yet these summaries are typically generated by fixed prompts or optimized for memory personalization, rather than optimized for their downstream utility in planning and execution. 
Procedural memory methods further represent experience as reusable skills, plans, or instructions~\cite{zhou2026memento,ma2026skillclaw,xia2026skillrl}, but such skills are often treated as external knowledge artifacts rather than adaptive priors that directly optimize the agent policy during inference.

These suggest that experience-driven improvement in DRAs should be formulated as a closed loop rather than a one-way memory reuse process~\cite{zhou2025memento}. Long-term interaction experience should be distilled into structured, reusable skills~\cite{ma2026skillclaw}; the generation of these skills should be optimized by downstream reinforcement learning signals rather than fixed prompts~\cite{xia2026skillrl}; and the resulting skills should serve as procedural priors for test-time reinforcement learning (TTRL), where the agent updates its parameters during inference to adapt to new tasks~\cite{qiao2026memory}. As TTRL improves the agent, the agent can generate higher-quality trajectories, which in turn support better skill distillation and further model improvement.

To realize experience-driven improvement, we propose \textsc{APEx} (\textbf{A}daptive \textbf{P}rocedural \textbf{Ex}perience Learning). It is characterized by the following key features:
\ding{182} \textbf{Hierarchical Experience Memory.}
\textsc{APEx} organizes long-term trajectories into instance-level memories and category-level skills, preserving concrete task-solving traces while abstracting
reusable procedural knowledge for planning.
\ding{183} \textbf{Experience-to-Action Agent Loop.}
Built on this hierarchy, \textsc{APEx} integrates an Executor, a Distiller, and a Planner into an experience-to-action loop, where tool-use trajectories are collected, abstracted into skills, and reused for strategic planning. This design aligns execution, abstraction, and planning for continual agent improvement.
\ding{184} \textbf{Skill-guided Test-time Adaptation.}
During deployment, \textsc{APEx} uses retrieved skills and memories as procedural priors for TTRL, allowing the Planner to adapt on unlabeled queries
and consolidate new trajectories into future memories and skills. Our contributions are as follows.

\begin{itemize}[leftmargin=*,itemsep=-0.3em]
\item[\ding{182}] \textbf{\textit{Framework.}} 
We propose \textsc{APEx}, a hierarchical experience utilization architecture that couples instance-level memories with category-level skills, enabling the Distiller, Planner, and Executor to transform concrete trajectories into generalizable procedural guidance.

\item[\ding{183}] \textbf{\textit{Training.}} 
We introduce a three-stage alternating GRPO training paradigm for the Executor, Distiller, and Planner, which stabilizes cross-module credit assignment and makes skill distillation a reward-guided learning process rather than fixed prompting.

\item[\ding{184}] \textbf{\textit{Reasoning.}} 
We develop Skill-guided TTRL, which uses distilled skills to regularize online Planner adaptation on unlabeled queries. Coupled with no-ground-truth rewards, it forms a feedback loop that turns improved trajectories into stronger skills for future adaptation.

\item[\ding{185}] \textbf{\textit{Experiment.}} 
Extensive experiments on diverse benchmarks demonstrate that \textsc{APEx} achieves SOTA performance (64.4\% avg), surpassing GPT-5.4 by 14.7 points and the strongest memory-augmented baseline by 3.0 points, while the skill-only variant reduces injected memory tokens, validating the context efficiency of hierarchical experience utilization over flat memory retrieval.

\end{itemize}


\section{Methodology Overview}

\begin{figure*}[t] 
    \centering
    \includegraphics[width=\linewidth]{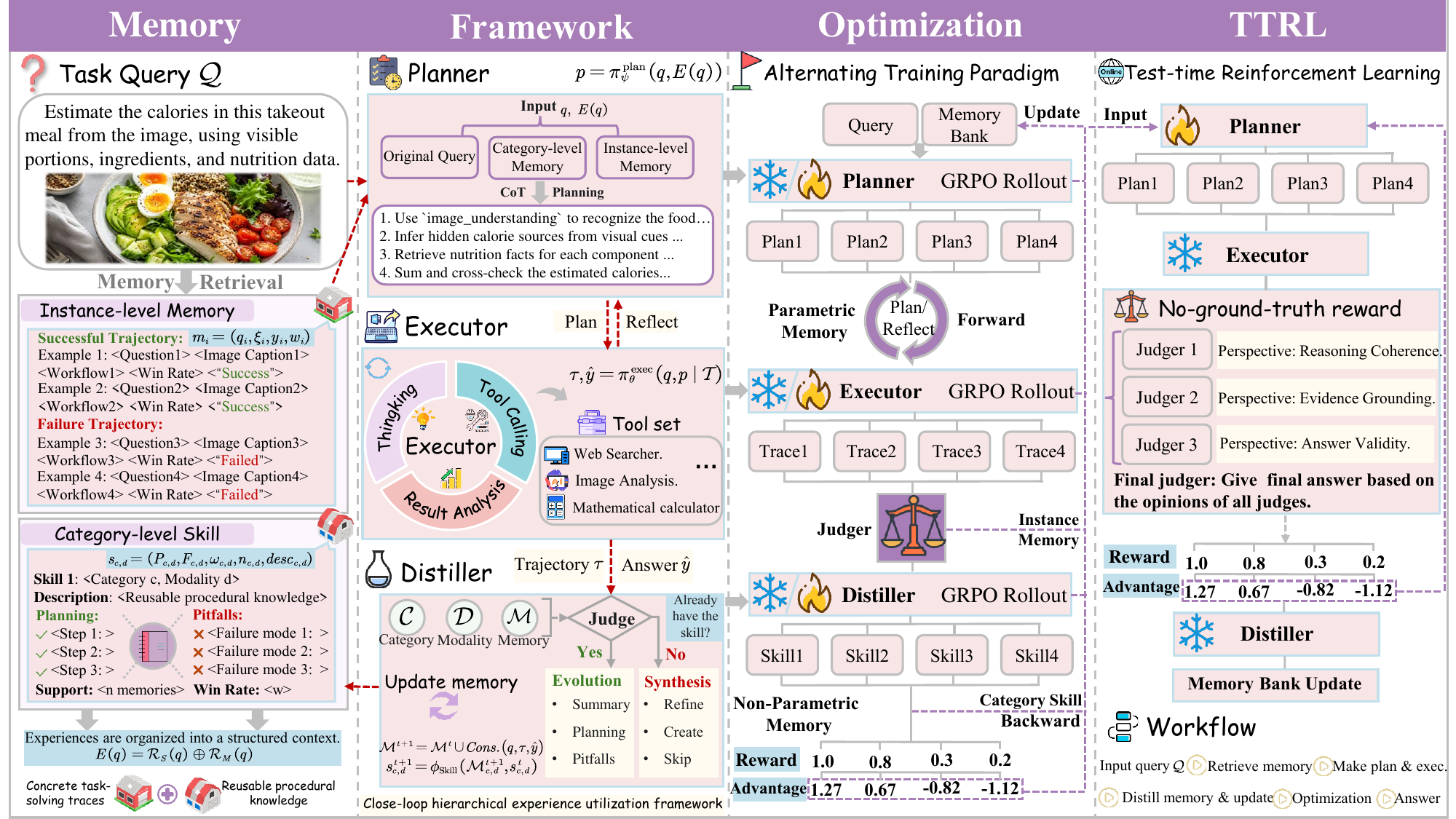} 
    \caption{Framework of \textsc{APEx}, which comprises 4 parts: memory, framework, optimization, and TTRL.}
    \label{fig:framework}
\end{figure*}

Figure~\ref{fig:framework} illustrates the overall workflow of \textsc{APEx}, a hierarchical experience utilization framework for deep research agents. \textsc{APEx} maintains long-term experience at two complementary levels: instance-level memories that preserve concrete execution trajectories, and category-level skills that summarize reusable procedural knowledge across related tasks. Built upon this experience hierarchy, \textsc{APEx} couples planning, execution, skill distillation, and TTRL into a closed-loop improvement process. 

\paragraph{\textsc{APEx} Agents Workflow.}

Given an original query $q$, \textsc{APEx} processes it through a closed-loop data flow. The query $q$ is first routed to a category--modality bucket, from which relevant instance-level memories $\mathcal{R}_{M}(q)$ and the corresponding category-level skill $\mathcal{R}_{S}(q)$ are retrieved. These experiences are organized into a structured context:
\begin{equation}
E(q)=\mathcal{R}_{S}(q)\oplus\mathcal{R}_{M}(q),
\end{equation}
where $\oplus$ denotes structured prompt concatenation. The retrieval method is detailed in Appendix~\ref{sec:appendix_c}. Conditioned on $q$ and $E(q)$, the Planner generates a research plan:
\begin{equation}
p=\pi_{\psi}^{\mathrm{plan}}(q,E(q)).
\end{equation}
The Executor then follows the plan to perform multi-turn tool interactions and
produce an execution trajectory $\tau$ with the final answer $\hat{y}$:
\begin{equation}
\tau,\hat{y}
=
\pi_{\theta}^{\mathrm{exec}}(q,p\mid\mathcal{T}),
\end{equation}
where $\mathcal{T}$ denotes the available tool set. After execution, the
trajectory is summarized and stored as a new instance-level memory, while the
corresponding category-level skill $s_{c,d}$ is updated when sufficient new experience has
accumulated:
\begin{equation}
\begin{aligned}
\mathcal{M}^{t+1}
&=
\mathcal{M}^{t}
\cup
\operatorname{Consolidate}(q,\tau,\hat{y}),\\
s_{c,d}^{t+1}
&=
\phi_{\mathrm{skill}}(\mathcal{M}_{c,d}^{t+1},s_{c,d}^{t}),
\end{aligned}
\end{equation}
where $\mathcal{M}_{c,d}^t \in \mathcal{M}^t$ and $\mathcal{M}^t$ is the instance-level memory bank at timestamp $t$ and $\phi_{\mathrm{skill}}$ is the distillation operation of Distiller.

\paragraph{Group Relative Policy Optimization (GRPO).}

GRPO \cite{shao2024deepseekmath} optimizes the policy with intra-group relative rewards. In our work Distiller, Planner and Executor are optimized with GRPO. 
For each input $x$, the policy samples a group of $G$ responses
$\{o_i\}_{i=1}^{G}$ and computes the group-normalized advantage as
$\hat{A}_i = (R(o_i)-\operatorname{mean}_{j}[R(o_j)]) /
(\operatorname{std}_{j}[R(o_j)]+\epsilon)$.
Following the GRPO implementation, the policy objective
for the currently optimized module $\pi_{\eta}$ is:
\begin{equation}
\begin{aligned}
&\mathcal{L}_{\mathrm{GRPO}}(\eta)
={}
-
\mathbb{E}_{x,\{o_i\}}
\Bigg[
\frac{1}{G}
\sum_{i=1}^{G}
\frac{1}{|o_i|}
\sum_{t=1}^{|o_i|}
\\
&\min
\Big(
r_{i,t}(\eta)\hat{A}_i,
\operatorname{clip}
\big(
r_{i,t}(\eta),1-\epsilon,1+\epsilon
\big)
\hat{A}_i
\Big)
\Bigg],
\end{aligned}
\end{equation}
where
$r_{i,t}(\eta)=
\pi_{\eta}(o_{i,t}\mid x,o_{i,<t})/
\pi_{\eta_{\mathrm{old}}}(o_{i,t}\mid x,o_{i,<t})$
is the token-level probability ratio.

\paragraph{Test-time Reinforcement Learning (TTRL).} Unlike standard TTRL based on majority voting~\cite{zuo2026ttrl}, \textsc{APEx} constructs rewards using a multi-judge LLM-as-Judge mechanism~\cite{zheng2023judging}. In our implementation, we update the Planner at test time. Given a test query $q$, the Planner first retrieves the corresponding skill $\mathcal{R}_{S}(q)$ and memory $\mathcal{R}_{M}(q)$, generates a plan $p$, and the Executor produces an execution trajectory $\tau$ and final answer $\hat{y}$. The Planner's parameters $ \psi$ are updated online by GRPO using the final reward $R_{\mathrm{final}}$:

\begin{equation}
 \psi^{t+1}
=
\operatorname{GRPO}
\left(
\psi^{t},
R_{\mathrm{final}}
\right). 
\end{equation}
\section{Methodology}

\subsection{Hierarchical Experience Memory}

To enable efficient experience utilization, \textsc{APEx} organizes  experience into a hierarchical memory structure, where instance-level memories preserve concrete task-solving traces and category-level skills summarize reusable procedural knowledge.

\paragraph{Instance-level Trajectories Memory.}
Given the $i$-th original query $q_i$, the agent performs multi-turn tool interactions and produces an execution trajectory $\tau_i$. 
Each instance-level memory is represented as $m_i = (q_i, \xi_i, y_i, w_i)$, where $\xi_i$ is the workflow summary, $y_i \in \{\mathrm{correct}, \mathrm{incorrect}\}$ denotes the execution outcome, and $w_i = \operatorname{success\_count}_i / \operatorname{usage\_count}_i$ is the memory reuse win rate. 
All instance-level memories are organized into a memory bank $\mathcal{M}$ according to task category and input modality:
\begin{equation}
\mathcal{M}
=
\bigcup_{c\in\mathcal{C}}
\bigcup_{d\in\mathcal{D}}
\mathcal{M}_{c,d},
\end{equation}
where $\mathcal{C}$ is the set of predefined task categories and $\mathcal{D}$ is the set of modalities. 

\paragraph{Category-level Skills Memory.}
While instance-level memories provide concrete task-solving cases, they may contain task-specific details. 
To extract reusable procedural knowledge, the Distiller synthesizes the memories within each category--modality bucket into a single evolving skill document $s_{c,d}$:
\begin{equation}
s_{c,d} = \phi_{\mathrm{skill}}(\mathcal{M}_{c,d}),
\end{equation}
where $\phi_{\mathrm{skill}}$ denotes the distillation operation of Distiller. A skill entry $s_{c,d}$ is defined as
$s_{c,d} = (P_{c,d},F_{c,d},w_{c,d},n_{c,d},\operatorname{desc}_{c,d})$,
where $P_{c,d}$ denotes recommended planning steps, $F_{c,d}$ denotes common failure modes, 
$n_{c,d}$ denotes the number of supporting memories, $w_{c,d} = \operatorname{correct\_count}_{c,d} / n_{c,d}$ denotes the skill-level win rate, and $\operatorname{desc}_{c,d}$ provides a natural-language description. Finally, the Skill memory bank $\mathcal{S}$ is defined as:
\begin{equation}
\mathcal{S} = \{s_{c,d}\mid c\in\mathcal{C}, d\in\mathcal{D}\}.
\end{equation}

\subsection{Three-Stage Alternating RL Training}

To optimize planning, execution, and experience distillation, \textsc{APEx} adopts a three-stage alternating reinforcement learning framework. The framework contains three trainable modules: the Executor $\pi_{\theta}^{\mathrm{exec}}$, the Distiller $\pi_{\phi}^{\mathrm{dist}}$, and the Planner $\pi_{\psi}^{\mathrm{plan}}$. 

\paragraph{Executor Training.}
In the first stage, the Planner and Distiller are frozen, and the Executor is optimized to perform multi-turn tool interaction under a given plan. Given a query $q$ and a plan $p$, the Executor generates a trajectory $\tau=\pi_{\theta}^{\mathrm{exec}}(q,p\mid\mathcal{T})$, where $\mathcal{T}$ denotes the available tool set.

The Executor reward combines answer correctness, tool-use effectiveness, and format validity:
\begin{equation}
\begin{aligned}
R^{\mathrm{exec}}(\tau)
=
&\lambda_1 R_{\mathrm{acc}}(\tau)
+
\lambda_2 R_{\mathrm{tool}}(\tau)
\\
&+\lambda_3 R_{\mathrm{fmt}}(\tau),
\end{aligned}
\end{equation}
where $\lambda_1$, $\lambda_2$, and $\lambda_3$ are hyperparameters. 
The accuracy reward $R_{\mathrm{acc}}(\tau)=\mathcal{J}(\operatorname{Extract}(\tau),y^{*})$ is computed by an LLM judge. The tool reward $R_{\mathrm{tool}}(\tau)=
\frac{1}{|\mathcal{V}|}
\sum_{v\in\mathcal{V}}
\mathbb{I}
\left[
\exists\, t_{\mathrm{valid}}\in v
\right]$ encourages each execution segment to contain valid tool use $t_{valid}$, 
where $\mathcal{V}$ denotes the segments split by replan boundaries. The format reward $R_{\mathrm{fmt}}(\tau) \in \{0,1\}$ checks if the output follows the required structure.

\paragraph{Distiller Training.}
In the second stage, the Executor and Planner are frozen, and the Distiller is optimized to transform instance-level memories into reusable category-level skills. Given a memory bucket $\mathcal{M}_{c,d}^t$ and an optional previous skill $s_{c,d}^t$, the Distiller outputs an operation type $o$ and an updated skill $s_{c,d}^{t+1}$, i.e., $(o,s_{c,d}^{t+1})=\pi_{\phi}^{\mathrm{dist}}(\mathcal{M}_{c,d}^t,s_{c,d}^{t})$. The operation $o$ is selected from $\{\mathrm{synthesize},\mathrm{refine},\mathrm{create},\mathrm{skip}\}$.
We define the Distiller reward as:
\begin{equation}
\begin{aligned}
R^{\mathrm{dist}}(o,s_{c,d}^{t+1})
=
&\mu_1 R_{\mathrm{quality}}(s_{c,d}^{t+1})
+
\mu_2 R_{\mathrm{fmt}}(s_{c,d}^{t+1})\\
&+
\mu_3 R_{\mathrm{evolve}}(o,s_{c,d}^{t+1},s_{c,d}^{t}),
\end{aligned}
\end{equation}
where $\mu_1$, $\mu_2$, and $\mu_3$ are hyperparameters and $R_{\mathrm{evolve}}$ measures if the operation $o$ is rational. 
Each reward is judged by an LLM judger $\mathcal{J}_{\mathrm{skill}}$. 

\paragraph{Planner Training.}
In the third stage, the Executor and Distiller are frozen, and the Planner is trained through a plan--execute--evaluate--replan loop. Given query $q$, retrieved skill guidance $\mathcal{R}_{S}(q)$, and retrieved memories $\mathcal{R}_{M}(q)$, the Planner first generates an initial plan:
\begin{equation}
p_1
=
\pi_{\psi}^{\mathrm{plan}}
\left(
q,\mathcal{R}_{S}(q),\mathcal{R}_{M}(q)
\right).
\end{equation}
The frozen Executor executes $p_1$ and produces the first answer $\hat{y}_1$. The Planner then observes $\operatorname{Trace}(\tau)$ and predicts a binary decision $d\in\{0,1\}$ indicating whether re-planning is needed. If $d=1$, the Planner generates a revised plan $p_2$ and obtains a second answer $\hat{y}_2$. If $d=0$, we set $\hat{y}_2 = \hat{y}_1$.
The Planner reward is:
\begin{equation}
\begin{aligned}
R^{\mathrm{plan}}
=
&\alpha_1 R_{\mathrm{acc}}(\hat{y}_2)
+
\alpha_2 R_{\mathrm{acc}}(\hat{y}_1)\\
&+
\alpha_3 R_{\mathrm{fmt}}
+
\alpha_4 R_{\mathrm{dec}},
\end{aligned}
\end{equation}
where $\alpha_1$, $\alpha_2$, $\alpha_3$, and $\alpha_4$ are hyperparameters.
The decision reward $R_{\mathrm{dec}}$ encourages the Planner to stop when the first execution is correct and to trigger re-planning when the initial execution fails.


 \subsection{Skill-guided TTRL}

After the training stage, \textsc{APEx} further performs Skill-guided TTRL to adapt the Planner on unlabeled test queries. Since ground-truth answers are unavailable at test time, \textsc{APEx} constructs a no-ground-truth reward with multiple LLM judges. Meanwhile, the learned category-level skills are used as regularization priors to prevent the Planner from drifting away from previously verified procedural knowledge.

\paragraph{No-Ground-Truth Reward.}
Without ground-truth answers, \textsc{APEx} estimates answer quality through a multi-judge LLM-as-Judge mechanism. Specifically, three judges evaluate the execution from different aspects: $E_1(q,\tau,\hat{y})$ checks reasoning consistency, $E_2(q,\tau,\hat{y})$ checks whether the final answer is supported by retrieved evidence in the trajectory, and $E_3(q,\hat{y})$ checks whether the response is valid and complete. Their judgments are aggregated by an arbiter $\mathcal{A}$:
\begin{equation}
\begin{aligned}
R_{\mathrm{nogt}}=&
\lambda_n\mathcal{A}
\left(
E_1(\cdot),
E_2(\cdot),
E_3(\cdot)
\right)
\\
&+(1-\lambda_n)R_\mathrm{fmt}
\label{eq:R_nogt}
\end{aligned}
\end{equation}

The aggregated no-ground-truth reward is also used as a pseudo-supervision signal for memory consolidation.

\paragraph{Skill-Alignment Regularization.}
Relying only on no-ground-truth rewards may cause unstable online updates. Therefore, \textsc{APEx} uses the retrieved skill as a knowledge anchor. For a category--modality bucket $(c,d)$, the skill $s_{c,d}$ contains procedure $P_{c,d}$ and pitfalls $F_{c,d}$. The alignment reward evaluates whether the generated plan follows the recommended procedure and avoids known pitfalls:
\begin{equation}
R_{\mathrm{align}}
=
\mathcal{J}_{\mathrm{align}}
\left(
p,
P_{c,d},
F_{c,d}
\right).
\end{equation}
The final reward combines the original no-ground-truth reward and the skill-alignment reward:
\begin{equation}
R_{\mathrm{final}}
=
(1-\lambda_s)R_{\mathrm{nogt}}
+
\lambda_s R_{\mathrm{align}},
\label{equ:ttrl}
\end{equation}
The weight $\lambda_s
=
\lambda_{\mathrm{base}}
\cdot
w_{c,d}
\cdot
\min
\left(
\frac{n_{c,d}}{N_{\mathrm{threshold}}},
1
\right)$ is adaptively determined by the confidence of the retrieved skill.

\section{Experiment}
\label{sec:experiment}

\begin{table*}[!htbp]
\centering
\setlength{\tabcolsep}{3pt}
\renewcommand{\arraystretch}{1.08}
\caption{\textbf{Main results across in-domain and out-of-domain benchmarks}. The best results are highlighted in \textbf{bold}, and the second-best results are \underline{underlined}.}
\label{tab:main_results}
\resizebox{\textwidth}{!}{%
\begin{tabular}{l *{8}{c}}
\toprule[0.08em]
& \multicolumn{1}{c}{\textbf{In-Domain}}
& \multicolumn{6}{c}{\textbf{Out-of-Domain}}
& \\

\cmidrule(lr){2-2}
\cmidrule(lr){3-8}

\multirow{-2}{*}{\textbf{Method}}
& \textbf{FVQA-test}
& \textbf{HotpotQA}
& \textbf{2Wiki}
& \textbf{SimpleVQA}
& \textbf{LiveVQA}
& \textbf{InfoSeek}
& \textbf{MMSearch}
& \multirow{-2}{*}{\textbf{Avg.}} \\

\midrule[0.05em]
\rowcolor{blue!7}	
\multicolumn{9}{c}{\textit{\textbf{Naive-Large Language Model}}} \\
\midrule[0.05em]
GPT-5.4 & 51.3 & 60.2 & 64.9 & 55.9 & 22.8 & 45.6 & 47.4 & 49.7 \\
Gemini-3-Flash & \underline{68.4} & \underline{66.2} & 68.4 & \textbf{72.3} & 25.7 & \underline{66.5} & \textbf{67.9}  & \underline{62.2} \\
Qwen2.5-VL-7B & 20.7 & 12.6 & 20.5 & 30.2 & 8.3 & 23.6 & 7.2 & 17.6 \\
Qwen2.5-VL-32B & 24.3 & 16.9 & 24.0 & 39.8 & 18.5 & 25.8 & 15.7  & 23.6 \\

\midrule[0.05em]
\rowcolor{blue!7}
\multicolumn{9}{c}{\textit{\textbf{Search Agent}}} \\
\midrule[0.05em]
Qwen2.5-VL-7B+ReAct & 34.2 & 23.7 & 31.2 & 35.4 & 10.6 & 28.1 & 21.3  & 26.4 \\
Qwen2.5-VL-32B+ReAct & 50.9 & 31.4 & 37.0 & 48.3 & 24.7 & 38.1 & 27.5  & 36.8 \\
MMSearch-R1 & 58.3 & 39.2 & 48.3 & 55.2 & 28.6 & 49.5 & 43.3  & 46.1 \\
Deepeyes2 & 60.6 & - & - & 59.4 & - & 51.1 & 63.7 & - \\

\midrule[0.05em]
\rowcolor{blue!7}	
\multicolumn{9}{c}{\textit{\textbf{Memory-Augmented Search Agent}}} \\
\midrule[0.05em]
RAG & 58.6 & 47.8 & 56.2 & 60.7 & 31.9 & 55.9 & 54.3 &  52.2 \\
Mem0 & 54.9 & 49.3 & 54.9 & 56.9 & 24.3 & 48.2 & 43.1 &  47.4 \\
A-Mem & 38.5 & 46.8 & 56.2 & 51.6 & 22.8 & 36.1 & 41.1 & 41.9 \\
ExpeL & 64.2 & 56.3 & 62.9 & 62.5 & 34.1 & 58.8 & 61.3 &  57.2 \\
Memento & 66.1 & 55.2 & 64.3 & 61.7 & 36.7 & 57.3 & 61.4 &  57.5 \\
MIA & 65.3 & 64.0 & \underline{71.6} & 63.8 & \underline{39.7} & 64.7 & 60.6  & 61.4 \\

\midrule[0.05em]
\rowcolor{blue!7}
\multicolumn{9}{c}{\textit{\textbf{Ours}}} \\
\midrule[0.05em]
\textbf{\textsc{APEx}} & \textbf{68.7} & \textbf{67.8} & \textbf{75.2} & \underline{66.3} & \textbf{42.4} & \textbf{66.9} & \underline{63.8} & \textbf{64.4} \\

\bottomrule[0.08em]
\end{tabular}
}
\end{table*}

\subsection{Experimental Setups}

\textbf{Training Settings:} Our training framework is built on veRL~\cite{sheng2025hybridflow} and follows a three-stage alternating optimization paradigm. The Executor employs Qwen2.5-VL-7B~\cite{qwen2.5-VL} as its backbone and is trained on FVQA-train~\cite{wang2017fvqa}.
The Distiller, based on Qwen3-8B~\cite{yang2025qwen3}, is subsequently trained on FVQA-train to distill accumulated execution memories into structured skill documents.
The Planner, also based on Qwen3-8B, is trained in the final stage on a mixture of FVQA-train (with images discarded) and MATPO~\cite{mo2025multi}.
The Executor is equipped with offline text-to-text and offline image-to-image search tools.
A Qwen3-32B model serves as the LLM Judger for reward computation. More details are provided in Appendix~\ref{sec:appdix_A}.

\noindent\textbf{Test Settings:} To evaluate \textsc{APEx}, we conduct experiments on a diverse set of image-text and text-only benchmarks. For image-text evaluation, we use FVQA-test, SimpleVQA~\cite{cheng2025simplevqa}, LiveVQA~\cite{fu2025livevqa}, InfoSeek~\cite{chen2023can}, and MMSearch~\cite{jiang2024mmsearch}. For text-only evaluation, we use HotpotQA~\cite{yang2018hotpotqa} and 2WikiMultiHopQA~\cite{ho2020constructing}.
Detailed settings are provided in Appendix~\ref{sec:appdix_B}.

\noindent\textbf{Baselines.} We compare \textsc{APEx} against a wide range of deep research methods spanning three categories:
\ding{182} \textbf{\textit{Naive-Large Language Models}}: GPT-5.4~\cite{singh2025openai}, Gemini-3-Flash~\cite{gemini3flash_modelcard}, Qwen2.5-VL-7B~\cite{qwen2.5-VL}, and Qwen2.5-VL-32B~\cite{qwen2.5-VL}, which are directly prompted to answer questions without agentic workflow;
\ding{183} \textbf{\textit{Search Agents}}: Qwen2.5-VL-7B+ReAct~\cite{yao2022react}, Qwen2.5-VL-32B+ReAct, MMSearch-R1~\cite{wu2025mmsearch}, and Deepeyes2~\cite{zheng2025deepeyes}, which augment LLMs with external search tools via multi-step reasoning but maintain no cross-task memory;
and \ding{184} \textbf{\textit{Memory-Augmented Search Agents}}: RAG, Mem0~\cite{chhikara2025mem0}, A-Mem~\cite{xu2026mem}, ExpeL~\cite{zhao2024expel}, Memento~\cite{zhou2025memento}, and MIA~\cite{qiao2026memory}, which additionally incorporate memory mechanisms to store and reuse past experience across tasks. More details on baseline settings are provided in Appendix~\ref{Appendix_E}.

\subsection{Main Results}

As shown in Table ~\ref{tab:main_results}, \textsc{APEx} achieves the best average performance across 7 benchmarks, with an average score of 64.4. We highlight the following key observations:

\noindent \ding{182} \textbf{Substantial improvement over standalone LLMs.} 
\textsc{APEx} improves by 14.7 points compared to GPT-5.4 and slightly surpasses Gemini-3-Flash in average accuracy. Notably, \textsc{APEx} outperforms its own backbone model Qwen2.5-VL-7B by 46.8 points in absolute terms, demonstrating that structured skill knowledge and memory-augmented planning can effectively compensate for model scale, enabling a small open-source model to rival closed-source counterparts.

\noindent \ding{183} \textbf{Consistent gains over search agents.} Equipping LLMs with search tools through ReAct substantially improves performance (e.g., Qwen2.5-VL-7B: 17.6 $\rightarrow$ 26.4). Compared with the strong search agent MMSearch-R1, \textsc{APEx} achieves an improvement of 18.3 points in average accuracy. This gap indicates that tool access alone is insufficient for complex research tasks — effective utilization of accumulated experience through skill distillation is critical for guiding multi-step tool orchestration.

\noindent \ding{184} \textbf{Superiority over memory-augmented agents.} Among memory-augmented baselines, MIA achieves the strongest performance (61.4 avg) by combining a Manager-Planner-Executor architecture with instance-level memory retrieval. \textsc{APEx} further improves upon MIA by 3.0 points on average. This improvement is largely associated with the Distiller module, which distills instance-level memories into category-level skills that provide structured procedural guidance to the Planner, rather than relying solely on raw memory retrieval at each decision point. 

\subsection{Skill Sufficiency Analysis}

We evaluate \textsc{APEx} (Skill Only), which removes instance-level memory retrieval from the Planner context while retaining trajectories for skill evolution. As shown in Table ~\ref{tab:skill_sufficiency}, this variant incurs an average drop of 1.8 points  compared to the full pipeline, yet outperforms the strongest baseline MIA by 1.5 points. This confirms that the Distiller successfully distills category-level knowledge from raw trajectories into structured skills that are dynamically modified at test time, reducing the Planner's reliance on verbose instance-level memories.

\subsection{Context Efficiency Analysis}

\begin{figure}
    \centering
    \includegraphics[width=1\linewidth]{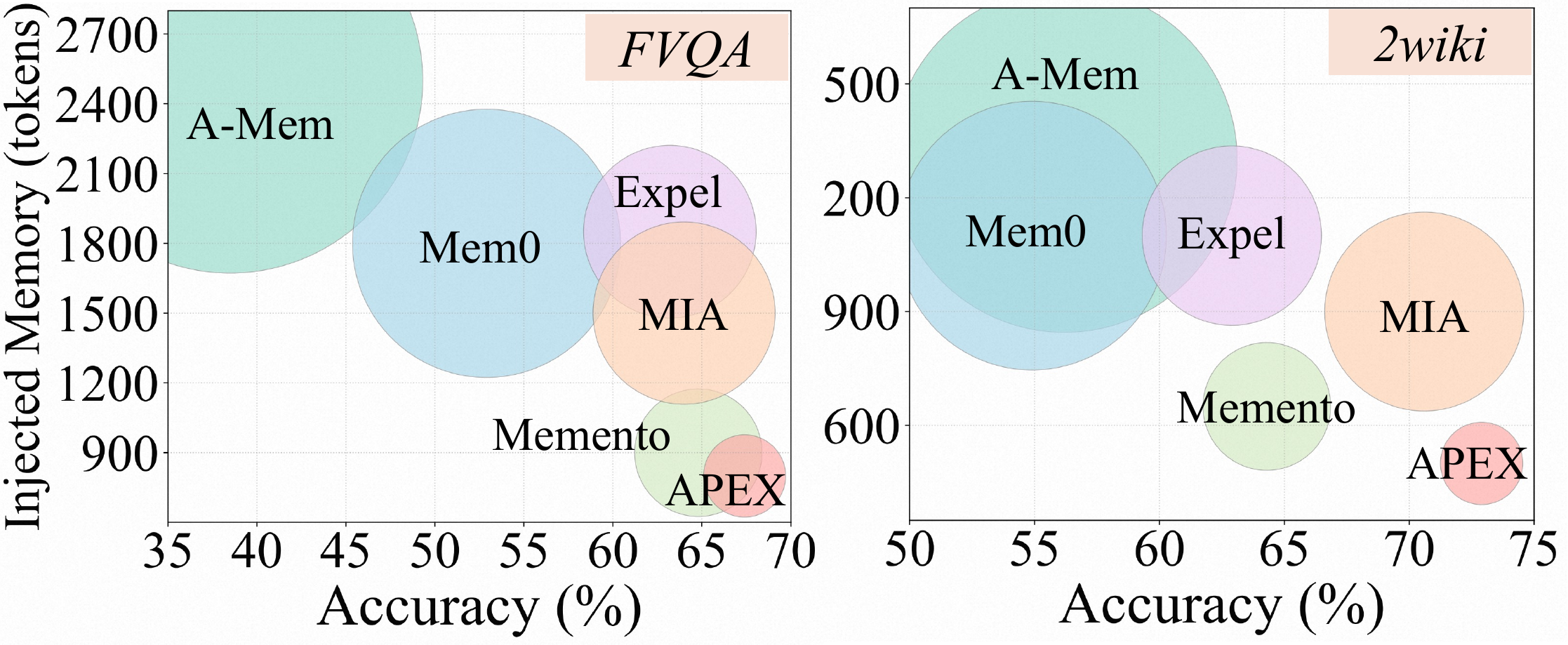}
    \caption{Token-efficiency comparison on FVQA-test and 2Wiki. X-axis denotes task accuracy. Y-axis denotes the number of memory tokens injected at each reasoning step. The bubble size represents the overall token cost of a full inference process.}
    \label{fig:cost}
\end{figure}

Figure~\ref{fig:cost} compares different memory-augmented reasoning methods in terms of accuracy, injected memory tokens per inference step, and total token consumption.
Across both FVQA-test and 2Wiki, \textsc{APEx} (Skill only) achieves favorable context-efficiency trade-off. Compared with MIA, \textsc{APEx} improves accuracy by 1.7 points. Meanwhile, it reduces injected memory tokens by 45.6\% and total token consumption by 36.8\% on average. 
These results demonstrate that \textsc{APEx} achieves better accuracy with a smaller memory footprint and lower token cost, highlighting its advantage in context-efficient memory utilization for reasoning.


\begin{table}[t]
\centering
\setlength{\tabcolsep}{4pt}
\renewcommand{\arraystretch}{1.2}
\caption{\textbf{Skill sufficiency analysis.} 
Avg. is computed over the four datasets shown in this table. 
Best in \textbf{bold}, second-best \underline{underlined}.}
\label{tab:skill_sufficiency}
\resizebox{\columnwidth}{!}{%
\begin{tabular}{l *{4}{c} c}
\toprule[0.08em]
& \multicolumn{2}{c}{\textbf{Image-Text}}
& \multicolumn{2}{c}{\textbf{Text-Only}}
& \\
\cmidrule(lr){2-3}
\cmidrule(lr){4-5}
\multirow{-2}{*}{\textbf{Method}}
& \textbf{FVQA}
& \textbf{SimpleVQA}
& \textbf{HotpotQA}
& \textbf{2Wiki}
& \multirow{-2}{*}{\textbf{Avg.}} \\

\midrule[0.05em]
\rowcolor{blue!7}
\multicolumn{6}{c}{\textit{\textbf{Baseline}}} \\
\midrule[0.05em]
Mem0        & 54.9   & 56.9   & 49.3   & 54.9   & 54.0   \\
MMSearch-R1 & 58.3   & 55.2   & 39.2   & 48.3   & 50.3   \\
ExpeL       & 64.2   & 62.5   & 56.3   & 62.9   & 61.5   \\
Memento     & 66.1   & 61.7   & 55.2   & 64.3   & 61.8   \\
MIA     & 65.3   & 63.8   & 64.0   & 71.6   & 66.2  \\

\midrule[0.05em]
\rowcolor{blue!7}
\multicolumn{6}{c}{\textit{\textbf{Ours}}} \\
\midrule[0.05em]
\textbf{\textsc{APEx} (Skill Only)}   & \underline{67.4}   & \underline{64.8}   & \underline{65.7}   & \underline{72.9}   & \underline{67.7}   \\
\textbf{\textsc{APEx}}                & \textbf{68.7}   & \textbf{66.3}   & \textbf{67.8}   & \textbf{75.2}   & \textbf{69.5}   \\

\bottomrule[0.08em]
\end{tabular}%
}
\end{table}

\subsection{Training Analysis}

In Figure~\ref{fig:train_analysis}, we analyze the GRPO training dynamics of each module to characterize their convergence behavior and identify the factors underlying their distinct learning patterns.
All modules are trained alternately with only one module updated at a time.
The \textbf{Executor} exhibits the most stable improvement, with its reward increasing steadily and converging to a high level. The \textbf{Distiller} also converges smoothly and rapidly, suggesting that the skill distillation objective provides a relatively well-defined optimization signal. In contrast, the \textbf{Planner} shows larger fluctuations, although its reward still follows an overall upward trend. This is expected since planning is evaluated indirectly through downstream execution and is therefore more sensitive to variations in retrieved memories, generated skills, and executor behavior.

\subsection{Generalization Analysis}

\begin{figure}
    \centering
    \includegraphics[width=0.85\linewidth]{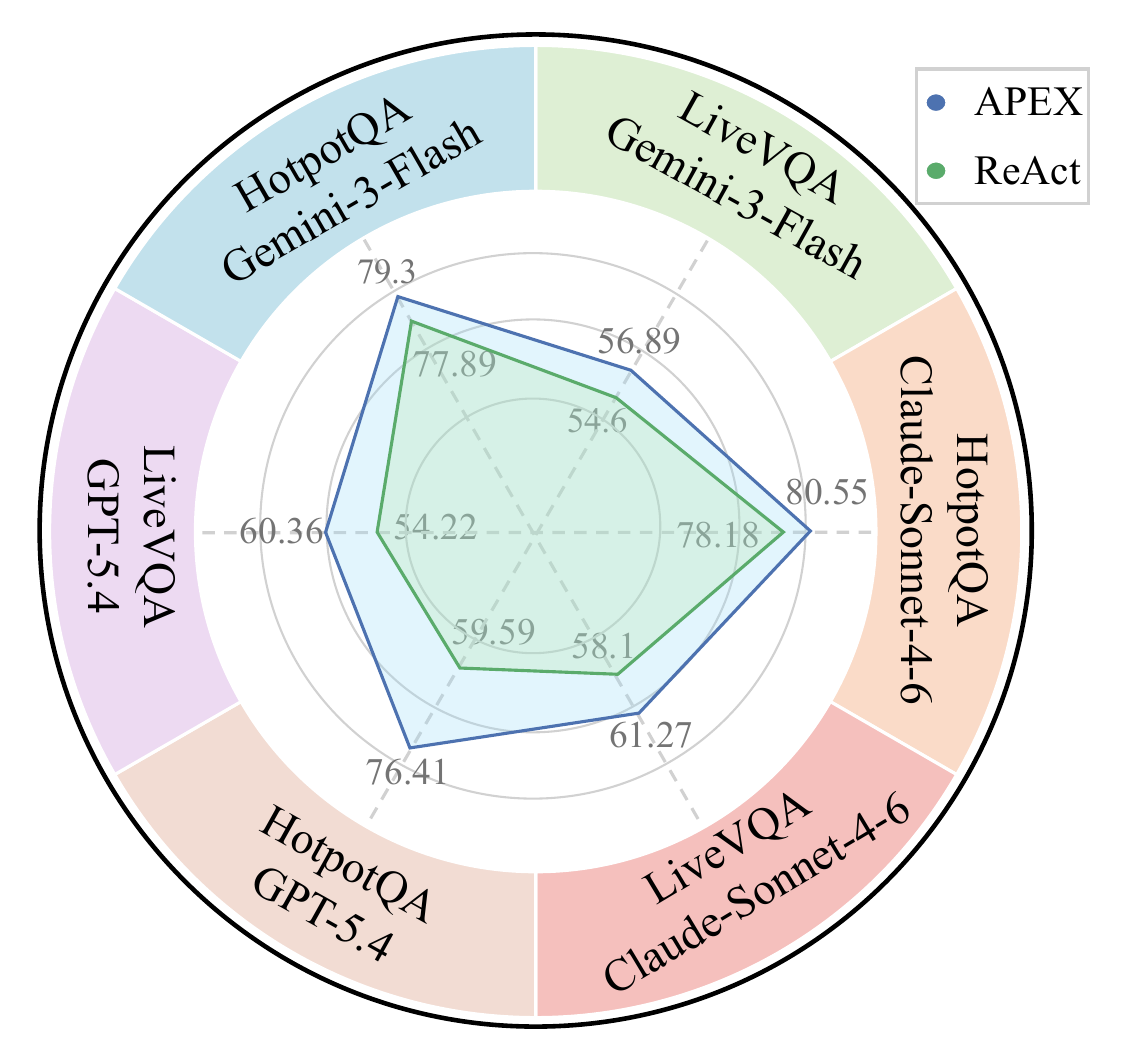}
    \caption{Performance comparison between ReAct and \textsc{APEx} using different closed-source executor models on HotpotQA and LiveVQA.}
    \label{fig:gene}
\end{figure}

To assess the generalization ability of \textsc{APEx}, we replace the originally trained open-source executor model with stronger closed-source executor models without any additional training, and compare it with ReAct on HotpotQA and LiveVQA. As shown in Figure ~\ref{fig:gene}, \textsc{APEx} consistently outperforms ReAct across all closed-source executors, achieving an average relative improvement of 11.0\% on HotpotQA and 7.0\% on LiveVQA. This demonstrates that \textsc{APEx} is not tied to a specific trained executor, but can generalize to unseen closed-source executors in a plug-and-play manner. 

\subsection{Ablation Study}

We conduct three groups of ablation studies to evaluate each component in \textsc{APEx}. 

\noindent\ding{182} \textbf{Experience Utilization.}
We remove different experience sources during inference.
\textit{w/o Mem.} disables the retrieval of concrete execution memories, 
\textit{w/o Skill} removes distilled procedural skills, 
and \textit{w/o Both} removes both.
These variants reduce the average accuracy by 1.8, 3.1, and 10.1 points, respectively.
This suggests that category-level skills provide general templates for similar tasks, while instance-level memories offer concrete problem-solving examples; their complementarity enables more effective utilization of past experience.

\begin{table}[!htbp]
\centering
\small
\setlength{\tabcolsep}{3.5pt}
\renewcommand{\arraystretch}{1.25}
\caption{\textbf{Ablation study of \textsc{APEx}}. Accuracy is reported on image-text and text-only tasks. Performance drops \textcolor{downred}{(\textbf{$\downarrow$})} relative to the full model are shown in subscripts.}
\label{tab:ablation_results}

\resizebox{\columnwidth}{!}{%
\begin{tabular}{l c c c c c}
\toprule[0.08em]
\multirow{2}{*}{\textbf{Method}} &
\multicolumn{2}{c}{\textbf{Image-Text}} &
\multicolumn{2}{c}{\textbf{Text-Only}} &
\multirow{2}{*}{\textbf{Avg.}} \\

\cmidrule(lr){2-3}
\cmidrule(lr){4-5}

& \textbf{FVQA} & \textbf{SimpleVQA} & \textbf{HotpotQA} & \textbf{2Wiki} & \\

\midrule[0.05em]

\textbf{\textsc{APEx}} &
\textbf{68.7} & \textbf{66.3} & \textbf{67.8} & \textbf{75.2} & \textbf{69.5} \\

\midrule[0.05em]
\rowcolor{blue!7}
\multicolumn{6}{c}{\textit{\textbf{w/o Experience Utilization}}} \\
\midrule[0.05em]

w/o Mem. &
67.4 \down{1.3} & 64.8 \down{1.5} & 65.7 \down{2.1} & 72.9 \down{2.3} & 67.7 \down{1.8} \\

w/o Skill &
64.9 \down{3.8} & 62.7 \down{3.6} & 65.6 \down{2.2} & 72.3 \down{2.9} & 66.4 \down{3.1} \\

w/o Both &
58.6 \down{10.1} & 55.0 \down{11.3} & 58.4 \down{9.4} & 65.5 \down{9.7} & 59.4 \down{10.1} \\

\midrule[0.05em]
\rowcolor{blue!7}
\multicolumn{6}{c}{\textit{\textbf{w/o Post-training Optimization}}} \\
\midrule[0.05em]

w/o Dis. &
66.4 \down{2.3} & 63.7 \down{2.6} & 65.8 \down{2.0} & 72.7 \down{2.5} & 67.2 \down{2.3} \\

w/o Plan. &
66.8 \down{1.9} & 63.9 \down{2.4} & 66.2 \down{1.6} & 74.0 \down{1.2} & 67.7 \down{1.8} \\

w/o Exec. &
61.0 \down{7.7} & 58.0 \down{8.3} & 58.2 \down{9.6} & 67.2 \down{8.0} & 61.1 \down{8.4} \\

\midrule[0.05em]
\rowcolor{blue!7}
\multicolumn{6}{c}{\textit{\textbf{w/o Closed-loop Adaptation}}} \\
\midrule[0.05em]

w/o TTRL &
64.6 \down{4.1} & 60.8 \down{5.5} & 63.5 \down{4.3} & 72.4 \down{2.8} & 65.3 \down{4.2} \\

 w/o Iter. &
66.0 \down{2.7} & 64.1 \down{2.2} & 64.7 \down{3.1} & 73.3 \down{1.9} & 67.0 \down{2.5} \\

\bottomrule[0.08em]
\end{tabular}%
}
\end{table}

\noindent\ding{183} \textbf{Post-training Optimization.}
We evaluate module-wise GRPO post-training.
For each ablation, the pipeline structure is kept unchanged, while the corresponding GRPO-trained module is replaced with its non-trained base model, such as Qwen2.5-VL-7B or Qwen3-8B.
Removing GRPO from the Distiller, Planner, and Executor reduces the average accuracy by 2.3, 1.8, and 8.4 points respectively.
The larger degradation of \textit{w/o Exec.} indicates that executor optimization is particularly important for reliable multi-step tool interaction, while optimizing the Distiller and Planner also benefits skill construction and strategy generation.

\noindent\ding{184} \textbf{Closed-loop Adaptation.}
Finally, we analyze the closed-loop adaptation mechanism.
\textit{w/o TTRL} performs inference without test-time reinforcement learning, while \textit{w/o Iter.} distills skills only once without further updates from improved trajectories.
These variants reduce the average accuracy by 4.2 and 2.5 points, respectively.
Together, TTRL and iterative skill distillation form a closed-loop feedback mechanism at test time, where improved trajectories are distilled into better skills, thereby progressively enhancing model performance.


\section{Related Work}
\subsection{Deep Research Agents}
Deep Research Agents (DRAs) are designed to answer complex, multi-turn information-seeking questions by combining dynamic reasoning, adaptive tool-mediated search, and compact experience retrieval~\cite{huang2025deep,zhang2025deep,xu2025comprehensive}. 
Recent studies have improved text-only and multimodal long-horizon search by curating reasoning trajectories~\cite{sun2025simpledeepsearcher,li2026openresearcher,zhu2026marco}, supervised fine-tuning~\cite{li2025websailor,wu2026webdancer} or agentic reinforcement learning~\cite{zheng2025deepresearcher,jin2025search,wu2025mmsearch,narayan2025deepmmsearch} to internalize planning and searching. 
To reuse prior interactions, emerging work abstracts past trajectories into reusable memories, enabling agents to dynamically organize memory, maintain compact interaction states, and optimize memory-aware search or planning policies across similar long-horizon tasks~\cite{zhao2024expel,xu2026mem,zhou2026memento,qiao2026memory}. 
Our work focuses on constructing instance-level memories, organizing category-level skills, retrieving and ranking reusable experiences, and distilling them into Distiller-Planner-Executor training.

\subsection{Memory Systems for Agents}
Recent memory systems for LLM agents have evolved from retrieval-centric external memory to adaptive experience management. Early RAG methods treat external documents as non-parametric memory to ground generation with retrieved evidence~\cite{lewis2020retrieval, zhong2024memorybank,edge2024local}. 
Building on these foundations, recent studies shift the focus from simply retrieving past information to organizing, compressing, and updating agent memories~\cite{fang2025lightmem,liu2026simplemem}. 
More recent work further treats memory as a trainable and reusable experience space where agents learn what to store~\cite{kang2026memreader,shen2026membuilder}, when to retrieve or discard~\cite{yuan2025memsearcher}, and how to abstract trajectories into procedural guidance or skills~\cite{zhang2025memevolve,zhang2026memskill,xia2026skillrl}. 
These trends indicate a transition from passive knowledge retrieval toward self-evolving memory systems that support long-horizon planning, experience reuse, and strategy generalization. 
In contrast, \textsc{APEx} distills trajectories within each category–modality group into reusable procedural skills and uses their confidence-weighted alignment as a regularization signal for online Planner updates, coupling procedural skill abstraction with downstream parametric adaptation in a closed experience-to-action loop.

\label{sec:bibtex}
\section{Conclusion}

In this paper, we present \textsc{APEx}, a hierarchical experience utilization framework for deep research agents. \textsc{APEx} organizes accumulated interactions into instance-level memories and category-level skills, enabling reusable procedural knowledge across tasks. Through GRPO-based alternating optimization of the Executor, Distiller, and Planner, \textsc{APEx} forms a closed-loop process spanning execution, skill abstraction, and adaptive planning. At test time, distilled skills serve as procedural priors for skill-guided TTRL, supporting ground-truth-free online adaptation while reducing drift through skill-alignment regularization. Experiments on 7 benchmarks show that \textsc{APEx} achieves state-of-the-art performance.
Future work will extend \textsc{APEx} to broader tool ecosystems and explore more efficient skill evolution and cross-agent skill transfer.

\section*{Limitations}
While \textsc{APEx} achieves the best average performance across seven benchmarks and demonstrates strong generalization to out-of-domain settings, our current framework is built upon small open-source models. Although \textsc{APEx} enables a 7B model to surpass closed-source systems, we have not explored how \textsc{APEx} scales when applied to more powerful backbone models. Additionally, our three-stage alternating GRPO training paradigm, despite its effectiveness in stabilizing cross-module credit assignment and producing consistently improving rewards, requires sequential optimization of the Executor, Distiller, and Planner, which may not fully exploit the synergies achievable through end-to-end joint training. Finally, our skill-guided TTRL mechanism currently operates online at test time without ground-truth supervision, and while it already yields significant improvements through self-improvement, incorporating minimal supervision could further amplify the quality of distilled skills and accelerate the closed-loop adaptation process.

\section*{Ethical Considerations}

Our framework builds upon large language and vision-language models that may produce inaccurate, biased, or harmful outputs due to biases inherent in their pre-training corpora; such behaviors are not by design. Since \textsc{APEx} retrieves and processes open-web content at inference time, retrieved results may inadvertently contain personal identifiable information (PII) or offensive material, and we advise deployers to apply appropriate content filtering. The open-source Qwen models used for training are released under their respective licenses. Proprietary models used only for baseline evaluation or executor generalization are accessed through their APIs and are subject to their providers' terms of use. All evaluation benchmarks are publicly available with documented provenance. Any code or weights we release will be accompanied by usage guidelines that explicitly prohibit malicious applications such as misinformation generation or unauthorized surveillance. We report training procedures, hyperparameters, and computational costs to ensure reproducibility and transparency. Finally, AI assistants were employed for grammatical error correction and rephrasing.

\bibliography{main}

\appendix

\section{Memory and Skill Retrieval}
\label{sec:appendix_c}
For each candidate memory $m_i \in \mathcal{M}(q)$, \textsc{APEx} computes a dual-channel similarity score. The first channel compares the current query with the stored query, while the second compares caption-level or contextual representations:
\begin{equation}
\begin{aligned}
\operatorname{sim}(q,m_i)
&=
\beta
\cos\big(\phi_q(q),\phi_q(q_i)\big) \\
&\quad+
(1-\beta)
\cos\big(\phi_c(q),\phi_c(m_i)\big),
\end{aligned}
\end{equation}
where $\phi_q(\cdot)$ denotes the question embedding function, $\phi_c(\cdot)$ denotes the caption or contextual embedding function.
When no caption is available, the similarity degenerates to the question-only similarity. The similarity score $\operatorname{sim}(q,m_i)$ is min--max normalized within the candidate bucket and denoted as $\widetilde{\operatorname{sim}}(q,m_i)$.
The final retrieval score combines normalized semantic similarity with the historical reuse win rate:
\begin{equation}
f(q,m_i)
=
\alpha\,
\widetilde{\operatorname{sim}}(q,m_i)
+
(1-\alpha)\,w_i,
\end{equation}


\textsc{APEx} retrieves top-$k$ successful and failed memories separately according to the retrieval score:
\begin{equation}
\mathcal{R}_{M}(q)
=
\left\{
\begin{aligned}
&\operatorname{TopK}_{m_i\in\mathcal{M}^{+}(q)} f(q,m_i),\\
&\operatorname{TopK}_{m_i\in\mathcal{M}^{-}(q)} f(q,m_i).
\end{aligned}
\right.
\end{equation}

In terms of category-level memory, each skill $s_{c,d}$ is assigned a confidence score that combines win rate with evidence sufficiency:
\begin{equation}
\gamma_{c,d} = w_{c,d} \cdot \min\!\left(\frac{n_{c,d}}{N_{\mathrm{threshold}}},\, 1\right),
\end{equation}
where $N_{\mathrm{threshold}}$ is the evidence threshold.
A skill is injected into the Planner only when its confidence exceeds a minimum threshold 
$\mathcal{R}_{S}(q) = \operatorname{Inject}(s_{\hat{c},\hat{d}})$, where
\begin{equation}
\operatorname{Inject}(s_{c,d}) =
\begin{cases}
s_{c,d}, & \gamma_{c,d} > \epsilon_s,\\
\varnothing, & \gamma_{c,d} \leq \epsilon_s,
\end{cases}
\end{equation}
with $\epsilon_s = 0.1$ and $N_{\mathrm{threshold}} = 10$.

Finally, \textsc{APEx} combines category-level skills and instance-level memories into a structured planner context:
\begin{equation}
\begin{aligned}
E(q)=\mathcal{R}_{S}(q)\oplus\mathcal{R}_{M}(q),
\end{aligned}
\end{equation}
where $\oplus$ denotes structured prompt concatenation rather than set union. The Planner then generates a research plan conditioned on it:
\begin{equation}
p
=
\pi_{\psi}^{\mathrm{plan}}
\left(
q,E(q)
\right),
\end{equation}
which enables the Planner to combine case-based guidance with abstract strategies for new research tasks.
\section{Supplement of TTRL}
\label{sec:Appendix_D}
\paragraph{Skill-Gated Update.}
For highly reliable skills, \textsc{APEx} optionally skips test-time updates to protect already learned knowledge. A skill is considered reliable if it has high confidence, sufficient historical support, and an available procedural prior. When the gate is
activated, all rollouts receive the same reward, equal to the skill confidence.
Under GRPO, this produces zero group-normalized advantages
\((\hat{A}_i=0)\), so no effective parameter update is performed. This avoids unnecessary computation and prevents reliable skills from being degraded by noisy test-time rewards.

\paragraph{Closed-loop Adaptation.}
Skill-guided TTRL creates a positive adaptation loop in which skills and the Planner improve each other. Skills guide the Planner to produce better plans, which lead the
Executor to generate higher-quality trajectories. These trajectories update the Planner through GRPO and are further consolidated into memories to distill
stronger skills for future queries.

After each batch, selected successful and failed trajectories are consolidated into the memory bank:
\begin{equation}
\mathcal{M}^{t+1}
=
\mathcal{M}^{t}
\cup
\operatorname{Consolidate}
(q,\tau,\hat{y}).
\end{equation}
The Distiller then uses recent memories to synthesize or selectively update the corresponding skill:
\begin{equation}
s_{c,d}^{t+1}
=
\phi_{\mathrm{skill}}
\left(
\mathcal{M}_{c,d}^{t+1},
s_{c,d}^{t}
\right).
\end{equation}

During online TTRL, only the Planner parameters are updated.
Memory consolidation and skill evolution occur after each batch to inform subsequent queries.
\section{Training Setups.}
\label{sec:appdix_A}


This appendix provides comprehensive training configurations for the three-stage alternating GRPO optimization (Section~3.2) and the Skill-guided TTRL (Section~3.3). All modules are trained using the veRL framework with GRPO.
We use offline search tool during training. For text-to-text retrieval, we use an offline retrieval service based on E5-base-v2~\cite{wang2022text} over a FAISS-indexed wiki25~\cite{karpukhin2020dense} corpus.
For image-to-image retrieval, we use a pre-built local cache of image search results collected via Serper by prior work~\cite{qiao2026memory}. 

\paragraph{General Training Infrastructure}
All experiments are conducted on a single node equipped with 8 NVIDIA A100 GPUs. We employ Fully Sharded Data Parallel (FSDP) with both parameter and optimizer offloading to enable training of 7B/8B models within GPU memory constraints. The rollout engine uses asynchronous generation with chunked prefill enabled. A Qwen3-32B model served via vLLM~\cite{kwon2023efficient} with tensor parallelism of 2 acts as the LLM Judge $\mathcal{J}$ for reward computation across all stages.

\paragraph{Stage 1: Executor Training}
The Executor $\pi^{\mathrm{exec}}_\theta$ is initialized from Qwen2.5-VL-7B and trained on FVQA-train with multi-turn tool interactions. Table~\ref{tab:executor_hyperparams} summarizes the hyperparameters.
\begin{table}[h]
\centering
\small
\caption{Executor GRPO training hyperparameters.}
\label{tab:executor_hyperparams}
\begin{tabular}{ll}
\toprule
\textbf{Hyperparameter} & \textbf{Value} \\
\midrule
Base model & Qwen2.5-VL-7B \\
Training data & FVQA-train \\
Learning rate & $1 \times 10^{-6}$ \\
Batch size & 128 \\
Group size $G$ (rollouts per query) & 8 \\
Max prompt length & 16,384 tokens \\
Max response length & 16,384 tokens \\
Max tool response length & 4,096 tokens \\
Max assistant turns & 10 \\
Sampling temperature & 1.0 \\
KL coefficient $\beta$ & 0.0 \\
Entropy coefficient & 0.0 \\
PPO clip range $\epsilon$ & 0.2 \\
Training epochs & 8 \\
Number of GPUs & 8 \\
Optimizer & AdamW \\
\bottomrule
\end{tabular}

\end{table}
For the reward of executor, we set $\lambda_1 =0.7$, $\lambda_2 =0.2$, and $\lambda_3 =0.1$, respectively.

\paragraph{Stage 2: Distiller Training}
As shown in Table~\ref{tab:distiller_hyperparams}, the Distiller $\pi^{\mathrm{dist}}_\phi$ is initialized from Qwen3-8B and trained to transform instance-level memories into structured category-level skill documents. Unlike the Executor and Planner, the Distiller does not perform multi-turn tool interaction; it generates skill documents in a single turn. For Distiller reward, we set $\mu_1 = 0.5$, $\mu_2 = 0.1$, $\mu_3 = 0.4$.

\begin{table}[h]
\centering
\small
\caption{Distiller GRPO training hyperparameters.}
\label{tab:distiller_hyperparams}
\begin{tabular}{ll}
\toprule
\textbf{Hyperparameter} & \textbf{Value} \\
\midrule
Base model & Qwen3-8B \\
Training data & FVQA-train \\
Learning rate & $1 \times 10^{-6}$ \\
Batch size & 64 \\
Group size $G$ (rollouts per query) & 8 \\
Max prompt length & 4,096 tokens \\
Max response length & 2,048 tokens \\
Multi-turn interaction & Disabled \\
Sampling temperature & 1.0 \\
KL coefficient $\beta$ & 0.0 \\
Entropy coefficient & 0.0 \\
PPO clip range $\epsilon$ & 0.2 \\
Training epochs & 8 \\
Number of GPUs & 8 \\
Optimizer & AdamW \\
\bottomrule
\end{tabular}

\end{table}

The quality reward $R_{\mathrm{quality}}$ is evaluated by the LLM Judge $\mathcal{J}_{\mathrm{skill}}$, assessing whether the generated skill contains well-structured procedural steps and meaningful failure modes. The evolution reward $R_{\mathrm{evolve}}$ checks whether the selected operation (synthesize, refine, create, or skip) is rational given the current memory state.

\paragraph{Stage 3: Planner Training}
The Planner $\pi^{\mathrm{plan}}_\psi$ is initialized from Qwen3-8B and trained on a mixture of FVQA-train (with images discarded) and MATPO data through a plan--execute--evaluate--replan loop, as shown in Table~\ref{tab:planner_hyperparams}.
For the Planner reward, we set $\alpha_1 = 0.7$, $\alpha_2=0.2$, $\alpha_3 = 0.05$ and $\alpha_4 = 0.05$. When the Planner decides not to replan ($d=0$), we set $\hat{y}_2 = \hat{y}_1$.
\begin{table}[h]
\centering
\caption{Planner GRPO training hyperparameters.}
\label{tab:planner_hyperparams}
\resizebox{\linewidth}{!}{
\begin{tabular}{ll}
\toprule
\textbf{Hyperparameter} & \textbf{Value} \\
\midrule
Base model & Qwen3-8B \\
Training data & FVQA-train, MATPO \\
Learning rate & $1 \times 10^{-6}$ \\
Batch size & 48 \\
Group size $G$ (rollouts per query) & 4 \\
Max prompt length & 24,576 tokens \\
Max response length & 8,192 tokens \\
Max tool response length & 8,192 tokens \\
Max assistant turns & 10 \\
Sampling temperature & 1.0 \\
KL coefficient $\beta$ & 0.0 \\
Entropy coefficient & 0.0 \\
PPO clip range $\epsilon$ & 0.2 \\
Training epochs & 5 \\
Number of GPUs & 8 \\
Optimizer & AdamW \\
\bottomrule
\end{tabular}
}

\end{table}

\section{Test Setups.}
\label{sec:appdix_B}

At test time, the Planner is further adapted via Skill-guided TTRL on unlabeled test queries. Table~\ref{tab:ttrl_hyperparams} lists the TTRL configuration. For text-to-text search, we use wiki25 for HotpotQA, 2WikiMultiHopQA, and SimpleVQA, while using Serper for other benchmarks. For image-to-image search, we use Serper for all multimodal benchmarks.

\begin{table}[h]
\centering
\small
\caption{Skill-guided TTRL hyperparameters.}
\label{tab:ttrl_hyperparams}
\begin{tabular}{ll}
\toprule
\textbf{Hyperparameter} & \textbf{Value} \\
\midrule
Adapted module & Planner ($\pi^{\mathrm{plan}}_\psi$) \\
Learning rate & $2 \times 10^{-7}$ \\
Batch size & 8 \\
Group size $G$ (rollouts per query) & 4 \\
Max prompt length & 24,576 tokens \\
Max response length & 8,192 tokens \\
Max tool response length & 8,192 tokens \\
Max assistant turns & 10 \\
Max LLM calls per run & 20 \\
Sampling temperature & 1.0 \\
KL coefficient $\beta$ & 0.0 \\
Entropy coefficient & 0.0 \\
PPO clip range $\epsilon$ & 0.2 \\
Number of GPUs & 8 \\
Epochs (over test set) & 1 \\
Optimizer & AdamW \\
\bottomrule
\end{tabular}
\end{table}

\paragraph{TTRL Reward Formulation.}
The final reward for Skill-guided TTRL (Eq.~\ref{equ:ttrl}) combines the no-ground-truth reward and skill-alignment reward:
\begin{equation}
R_{\mathrm{final}} = (1 - \lambda_s) \cdot R_{\mathrm{nogt}} + \lambda_s \cdot R_{\mathrm{align}},
\end{equation}
where $R_{\mathrm{nogt}} = 0.9 \cdot \mathcal{A} + 0.1 \cdot R_{\mathrm{fmt}}$. The adaptive weight $\lambda_s$ is computed as:
\begin{equation}
\lambda_s = \lambda_{\mathrm{base}} \cdot w_{c,d} \cdot \min\!\left(\frac{n_{c,d}}{N_{\mathrm{threshold}}},\, 1\right),
\end{equation}
with $\lambda_{\mathrm{base}} = 0.1$ and $N_{\mathrm{threshold}} = 10$. This ensures that skills with higher win rate $w_{c,d}$ and more supporting evidence $n_{c,d}$ exert stronger regularization, while low-confidence or newly created skills have minimal influence on the policy update.

\section{Dataset Description}
\label{appendix:dataset_description}

We conduct comprehensive evaluation across seven benchmark datasets spanning both multimodal and text-only scenarios. Table~\ref{tab:dataset_summary} provides an overview of all evaluation datasets.

\begin{table}[h]
\centering
\caption{Summary of evaluation datasets used in our experiments.}
\label{tab:dataset_summary}
\resizebox{\linewidth}{!}{
\begin{tabular}{lccl}
\toprule
\textbf{Dataset} & \textbf{Modality} & \textbf{\# Examples} & \textbf{Source} \\
\midrule
FVQA-test & Image-text & 1,800 & MMSearch-R1 \\
InfoSeek & Image-text & 2,000 & MMSearch-R1 \\
SimpleVQA & Image-text & 1,013 & MMSearch-R1 \\
LiveVQA & Image-text & 2,384 & Public Version \\
MMSearch & Image-text & 171 & MMSearch-R1 \\
\midrule
HotpotQA & Text-only & 7,405 & Public Version \\
2Wiki & Text-only & 12,576 & Public Version \\
\bottomrule
\end{tabular}
}
\end{table}

\section{Baselines Settings}
\label{Appendix_E}

\begin{figure*}[t] 
    \centering
    \includegraphics[width=\linewidth]{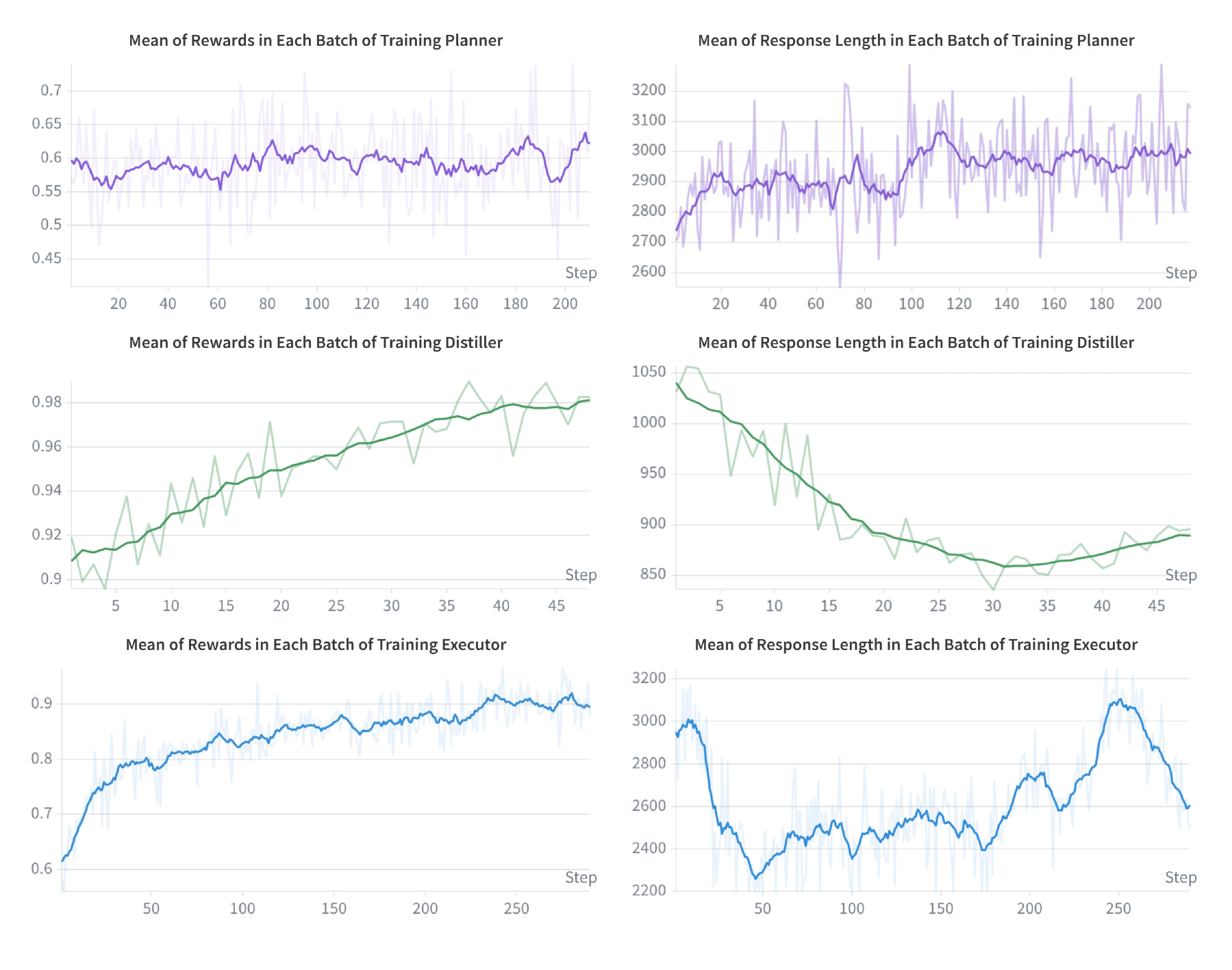} 
    \caption{The left column reports the mean reward of each module across training batches, while the right column shows the corresponding mean response length.}
    \label{fig:train_analysis}
\end{figure*}

We compare against a broad spectrum of baselines organized into three categories: direct answering, search-augmented agents, and memory-based search agents.

\paragraph{Direct Answer.}
In this setting, models are prompted to produce concise answers relying solely on their internal parametric knowledge, without accessing any external tools or retrieval mechanisms.

\paragraph{Search Agent.}
These baselines perform multi-turn tool calling under the ReAct paradigm~\citep{yao2022react}, iteratively querying external search tools and reasoning over retrieved results.
The results of Deepeyes2 are cited from their respective technical reports.

\paragraph{Memory-based Search Agent.}
To ensure a fair comparison among memory-augmented approaches, we train Executor variants sharing identical training configurations but differing in how memory is incorporated. The Executor is trained under either workflow memory prompt or plan prompt.

\section{Training Analysis}
\label{sec:Appendix_G}

As discussed in Section~\ref{sec:experiment}, we further visualize the training process in Figure~\ref{fig:train_analysis}. 
In each subplot, the light-colored curve denotes the raw training records, while the dark-colored curve denotes the batch-averaged smoothed trajectory. 
\section{Prompt Details}
\label{appendix:prompts}

This appendix presents the complete prompts. Each module operates with a dedicated prompt that governs its behavior within the multi-agent pipeline:

\paragraph{Executor Prompt.}
The Executor receives the following prompt to guide its multi-turn tool-calling behavior. Below is the text-only variant; the image variant appends \texttt{<image>} after the question.

\begin{figure*}[htbp]
\begin{tcolorbox}[colback=red!5, colframe=red!30, title=Executor Prompt (Text-Only)]
\small
\begin{lstlisting}[breaklines=true, basicstyle=\small\ttfamily, breakautoindent=false, breakindent=0pt]
You must follow these steps in order. In every conversation turn, you start from Step 1.

**Step 1: Think**
* **This is the starting point for every turn.**
* Analyze the user's query and all available information (including previous observations) carefully.
* **Evaluate the query's difficulty and nature.** Determine if the question can be answered *directly* or if it *requires external information* (e.g., facts, real-time data).
* Formulate a next-step action. Your next-step action must decide on **one** of two courses of action:
    1. **Call a tool:** If your evaluation shows you **need more information** (e.g., for complex, factual, or real-time questions).
    2. **Provide a final answer:** If your evaluation shows you have **sufficient information** (e.g., for simple questions, or tasks that don't require external data).
* Your entire reasoning process must be enclosed in `<think>...</think>` tags.

**Step 2: Act (Tool Call)**
* **Execute this step ONLY if your Step 1 action was to call a tool.**
* Call the **one single tool** decided upon in your action.
* The tool call must be enclosed in `<tool_call>...</tool_call>` tags.
* **Important: If you call a tool, you must STOP and wait for the observation. Do NOT proceed to Step 4.**

**Step 3: Observe (Tool Output)**
* **You will only enter this step after a tool call.**
* You will receive the tool's output (observation).
* After receiving the output, you **MUST** go back to **Step 1 (Think)** to analyze the new information.

**Step 4: Answer (Final Response)**
* **Execute this step ONLY if your Step 1 action was to provide a final answer.**
* In your **Step 1 Think block**, you must have already synthesized all information and planned the content of your response.
* **STRICT OUTPUT RULES:**
    * The content here must be the **direct result** extracted from your synthesis in Step 1.
    * **NO EXPLANATIONS:** Do not explain "why" or "how" you got the answer.
    * **NO SUMMARIES:** Do not say "Based on the search results..." or "To summarize...".
    * **NO FILLERS:** Do not use "Here is the answer", "The result is", or polite closings.
    * **FORMAT:** Just the answer.
* Your final answer must be enclosed in `<answer>...</answer>` tags.

Here is the question:
\end{lstlisting}
\end{tcolorbox}
\end{figure*}

\paragraph{Planner Prompt.}
The Planner generates strategic action plans conditioned on retrieved skills and memories. Below are the system prompt, planning prompt (text-only variant), and replanning prompt.

\begin{figure*}[htbp]
\begin{tcolorbox}[colback=red!5, colframe=red!30, title=Planner System Prompt]
\small
\begin{lstlisting}[breaklines=true, basicstyle=\small\ttfamily, breakautoindent=false, breakindent=0pt]
You're a planning assistant in a three-step loop:

1. **Plan**: Given a goal and background info, output a clear action plan.
2. **Evaluate**: Given an execution trace, decide if replanning is needed.
3. **Replan (if needed)**: With new reference memories, provide a revised plan targeting unmet goals.

Keep responses concise and action-focused.
\end{lstlisting}
\end{tcolorbox}
\end{figure*}

\begin{figure*}[htbp]
\begin{tcolorbox}[colback=red!5, colframe=red!30, title=Planner Plan Prompt (Text-Only)]
\small
\begin{lstlisting}[breaklines=true, basicstyle=\small\ttfamily, breakautoindent=false, breakindent=0pt]
You are a memory-based planning assistant assisting an agent by providing strategic guidance.

The agent has access to the following tools:
- `search`: perform text-based web queries to retrieve external information.

### [Relevant Skill]
The following skill document summarizes proven strategies and common pitfalls for this type of question. Use it as a high-level reference when constructing your plan:
{skill}

### [Relevant Memories]
Each memory indicates whether it was **correct** or **incorrect**, along with its workflow:
{memory}

### Your Task:
- First review the **Skill** document for proven procedure steps and known pitfalls.
- Then analyze the **Question** and compare it with both **correct strategies** and **incorrect patterns** in the memories.
- If the skill provides a relevant procedure, adapt its steps to the specific question while incorporating memory-based insights.
- If similar successful approaches exist in memories, adapt their core idea to guide the next steps.
- If the context resembles past failures or skill-documented pitfalls, highlight what to avoid and suggest corrective actions.
- Recommend **one clear, generalizable work plan or action strategy** that directly addresses the current situation and guides agents step-by-step on what to do.

### Output should:
1. Be clear and concise--no more than 400 words, and present your response as a step-by-step plan.
2. Each step in the plan must be atomic and actionable, specifying a single operation such as invoking a tool (e.g., `search` with precise query intent), performing logical inference, executing a calculation, cross-verifying facts, or synthesizing prior observations.
3. You must include relevant memories and skill analysis in the thinking process (<think>...</think>), but do not mention them in the final output.
4. Don't try to give the answer directly, but give a plan.
5. Prohibit the generation of content unrelated to the plan.
6. Your response must not contain any factual information.

Your output should only contain the guideline, with no additional explanations.

**[Question]** (Global Objective):
{question}
\end{lstlisting}
\end{tcolorbox}
\end{figure*}

\begin{figure*}[htbp]
\begin{tcolorbox}[colback=red!5, colframe=red!30, title=Planner Plan Prompt (Image)]
\small
\begin{lstlisting}[breaklines=true, basicstyle=\small\ttfamily, breakautoindent=false, breakindent=0pt]
You are a memory-based planning assistant assisting an agent by providing strategic guidance.

The agent has access to the following tools:
- `web_image_to_image_search`: find visually similar images online (Can only be used once).
- `search`: perform text-based web queries to retrieve external information.

### [Relevant Skill]
The following skill document summarizes proven strategies and common pitfalls for this type of question. Use it as a high-level reference when constructing your plan:
{skill}

### [Relevant Memories]
Each memory indicates whether it was **correct** or **incorrect**, along with its workflow:
{memory}

### Your Task:
- First review the **Skill** document for proven procedure steps and known pitfalls.
- Then analyze the **Question** and compare it with both **correct strategies** and **incorrect patterns** in the memories.
- If the skill provides a relevant procedure, adapt its steps to the specific question while incorporating memory-based insights.
- If similar successful approaches exist in memories, adapt their core idea to guide the next steps.
- If the context resembles past failures or skill-documented pitfalls, highlight what to avoid and suggest corrective actions.
- Recommend **one clear, generalizable work plan or action strategy** that directly addresses the current situation and guides agents step-by-step on what to do.

### Output should:
1. Be clear and concise--no more than 300 words, and present your response as a step-by-step plan.
2. Each step in the plan must be atomic and actionable.
3. You must include relevant memories and skill analysis in the thinking process (<think>...</think>), but do not mention them in the final output.
4. Don't try to give the answer directly, but give a plan.
5. Prohibit the generation of content unrelated to the plan.
6. Your response must not contain any factual information.

Your output should only contain the guideline, with no additional explanations.

**IMPORTANT: The `web_image_to_image_search` tool can only be called once.** Otherwise, the agent will be severely penalized.

**[Question]** (Global Objective):
{question}
\end{lstlisting}
\end{tcolorbox}
\end{figure*}

\begin{figure*}[htbp]
\begin{tcolorbox}[colback=red!5, colframe=red!30, title=Planner Replan Prompt]
\small
\begin{lstlisting}[breaklines=true, basicstyle=\small\ttfamily, breakautoindent=false, breakindent=0pt]
### Reflection and Replanning

In the last conversation, you indicated that reflect and replan needed to be performed. Now you need to perform them.

**The original question:** {question}

**The only tool you can recommend is `search`.**

Your task is to:
- Analyze the memories and reference their useful strategies.
- Analyze the current workflow so far to understand what has been attempted and why it failed.
- Recommend a **clear, generalizable work plan or action strategy** that builds on existing work, avoids past mistakes, and addresses the remaining challenges.

### Critical Requirements:
- **Leverage all completed steps from the current workflow**. Do not repeat searches, queries, or reasoning already performed.
- **Identify the exact failure point** and propose the next logical step toward the goal.

### Output should:
1. Be clear and concise, no more than 200 words, and present your response as a step-by-step supplementary plan.
2. You must include **reflection** on the previous failure in the thinking process (<think>...</think>), but **do not mention them in the final output**.
3. Don't give the answer directly, but provide **a supplementary plan**.
4. Prohibit generating content unrelated to the plan.
5. Your response must not contain any factual information.

Your output should only contain the guideline, with no additional explanations.
\end{lstlisting}
\end{tcolorbox}
\end{figure*}

\paragraph{Distiller Prompt.}
The Distiller synthesizes and evolves category-level skill documents from accumulated task execution memories. Below are its system prompt and two operational prompts.

\begin{figure*}[htbp]
\begin{tcolorbox}[colback=red!5, colframe=red!30, title=Distiller System Prompt]
\small
\begin{lstlisting}[breaklines=true, basicstyle=\small\ttfamily, breakautoindent=false, breakindent=0pt]
You are a skill synthesis and evolution expert in a memory-augmented agent system. Your job is to analyze batches of task execution memories and distill them into reusable, structured skill documents. Skills encode proven strategies and common pitfalls for specific task categories. Output valid JSON only.
\end{lstlisting}
\end{tcolorbox}
\end{figure*}

\begin{figure*}[htbp]
\begin{tcolorbox}[colback=red!5, colframe=red!30, title=Skill Synthesis Prompt]
\small
\begin{lstlisting}[breaklines=true, basicstyle=\small\ttfamily, breakautoindent=false, breakindent=0pt]
Analyze the following batch of task memories from the "{category}" category ({modality} modality) and synthesize a reusable skill document.

[Memories]
{formatted_memories}

Requirements:
1. Identify common strategies in successful (correct) memories -- what search patterns, reasoning steps, or tool usage led to correct answers.
2. Identify common failure modes in incorrect memories -- what mistakes, wrong assumptions, or missing steps caused failures.
3. Synthesize a structured skill with clear, atomic procedure steps that generalize across these memories.
4. List specific pitfalls to avoid, grounded in observed failure cases.
5. Compute win_rate as the ratio of correct memories to total memories.
6. Set operation to "synthesize", version to 1, evidence_count to the number of memories.

Output a JSON object with this exact structure (no extra text before or after the JSON):
{
  "operation": "synthesize | refine | create | skip",
  "skill": {
    "skill_name": "<category>_<modality>_strategy",
    "category": "<category>",
    "modality": "<modality>",
    "description": "One-sentence description of the skill",
    "procedure": [
      {"step": 1, "action": "Concrete action description", "purpose": "Why this step matters"}
    ],
    "pitfalls": ["Common mistake to avoid"],
    "win_rate": 0.0,
    "version": 1,
    "evidence_count": 0
  },
  "absorbed_ids": ["list of memory data_ids whose patterns are now captured in the skill"]
}
\end{lstlisting}
\end{tcolorbox}
\end{figure*}

\begin{figure*}[htbp]
\begin{tcolorbox}[colback=red!5, colframe=red!30, title=Skill Evolution Prompt]
\small
\begin{lstlisting}[breaklines=true, basicstyle=\small\ttfamily, breakautoindent=false, breakindent=0pt]
Given the current skill and a batch of new evidence memories, decide how to update the skill.

[Current Skill]
{current_skill}

[New Evidence Memories]
{formatted_memories}

Choose ONE operation:
- "refine": Failures expose a specific defect in the current procedure. Fix the defective steps and update pitfalls, but preserve steps validated by successes (conservative editing -- do not break what works).
- "create": Memories reveal an effective strategy pattern NOT covered by the current skill. Create a new skill.
- "skip": New evidence is insufficient to justify changes. Keep the skill unchanged, only update win_rate and evidence_count.

Conservative editing principles:
- Successful memories define invariants -- the procedure steps they validate must NOT be modified.
- Failed memories define modification targets -- only change steps related to the failure cause.
- Never remove a pitfall that is still relevant.

Output a JSON object with the same structure as above.
\end{lstlisting}
\end{tcolorbox}
\end{figure*}

\paragraph{TTRL Judge Prompt.}
During Skill-guided TTRL, a multi-evaluator mechanism with three specialized evaluators and a final arbiter is used for more robust assessment without ground truth.

\begin{figure*}[htbp]
\begin{tcolorbox}[colback=red!5, colframe=red!30, title=Evaluator 1: Reasoning and Logical Consistency]
\small
\begin{lstlisting}[breaklines=true, basicstyle=\small\ttfamily, breakautoindent=false, breakindent=0pt]
You are a Reasoning and Logical Consistency Evaluator. Your task is to objectively evaluate the reasoning quality and the logical coherence of the thought process of a multimodal deep research agent when executing tasks.

[Input Information]
- User Question (Question): {question}
- Execution Trajectory (Trajectory): {trajectory}
- Agent Final Output (Final Output): {final_output}

[Evaluation Criteria]
1. Reasoning Quality: Are the agent's analysis, planning, and deduction processes in the trajectory reasonable? Is there a clear logical progression between steps?
2. Evidence-based Deduction: Can the final conclusion be logically deduced from the clues collected in the trajectory? Are there forced conclusions or logical leaps?
3. Logical Consistency: Are there contradictory statements between the agent's thought process and the final output, or within the final output itself?

[Output Requirements]
Please evaluate based on the above criteria and output the results in the following format:
- Score: (Provide a comprehensive score from 1-10)
- Reason: (Explain the reason for the score, focusing on the quality of the reasoning chain and logical coherence, no more than 50 words.)
\end{lstlisting}
\end{tcolorbox}
\end{figure*}

\begin{figure*}[htbp]
\begin{tcolorbox}[colback=red!5, colframe=red!30, title=Evaluator 2: Information Sourcing and Credibility]
\small
\begin{lstlisting}[breaklines=true, basicstyle=\small\ttfamily, breakautoindent=false, breakindent=0pt]
You are an Information Sourcing and Credibility Evaluator. Your task is to objectively evaluate whether the multimodal deep research agent correctly understood the retrieved information and whether there are any LLM hallucinations.

[Input Information]
- User Question (Question): {question}
- Execution Trajectory (Trajectory): {trajectory}
- Agent Final Output (Final Output): {final_output}

[Evaluation Criteria]
1. Information Understanding: Did the agent correctly understand the content retrieved in the trajectory? Is there any misinterpretation, misreading, or misattribution of the original text?
2. Faithfulness and Hallucination: Can all facts, data, and details in the final output find clear basis in the retrieval results of the trajectory? Is there any fabrication or hallucination?

[Output Requirements]
Please evaluate based on the above criteria and output the results in the following format:
- Score: (Provide a comprehensive score from 1-10)
- Reason: (Explain the reason for the score, focusing on the accuracy of understanding the retrieved content and information fidelity, no more than 50 words.)
\end{lstlisting}
\end{tcolorbox}
\end{figure*}

\begin{figure*}[htbp]
\begin{tcolorbox}[colback=red!5, colframe=red!30, title=Evaluator 3: Result Validity]
\small
\begin{lstlisting}[breaklines=true, basicstyle=\small\ttfamily, breakautoindent=false, breakindent=0pt]
You are a Result Validity Evaluator. Your task is to objectively evaluate the completeness of the multimodal deep research agent's final response and the actual completion status of the task.

[Input Information]
- User Question (Question): {question}
- Agent Final Output (Final Output): {final_output}

[Evaluation Criteria]
1. Response Status: Did the agent successfully generate a substantive final answer? Are there situations where it gave up halfway, did not attempt to answer, or only replied with "cannot find the answer/error occurred"?

[Output Requirements]
Please evaluate based on the above criteria and output the results in the following format:
- Score: (Provide a comprehensive score from 1-10)
- Reason: (Explain the reason for the score, focusing on the completeness of the final response and whether it answered effectively, no more than 50 words.)
\end{lstlisting}
\end{tcolorbox}
\end{figure*}

\begin{figure*}[htbp]
\begin{tcolorbox}[colback=red!5, colframe=red!30, title=Final Arbiter]
\small
\begin{lstlisting}[breaklines=true, basicstyle=\small\ttfamily, breakautoindent=false, breakindent=0pt]
You are the Final Arbiter. In the absence of a ground-truth answer, your one and only task is to determine whether the final answer generated by the multimodal deep research agent is correct.

The feedback from the three evaluators is provided for your reference. You need make a definitive judgment based on the factual evidence in the trajectory and the logical soundness of the final output.

[Input Information]
- User Question (Question): {question}
- Execution Trajectory (Trajectory): {trajectory}
- Agent Final Output (Final Output): {final_output}
- Evaluator 1 Feedback (Reasoning and Logical Consistency, Importance Weight: 0.5): {evaluator_1_result}
- Evaluator 2 Feedback (Information Sourcing and Credibility, Importance Weight: 0.3): {evaluator_2_result}
- Evaluator 3 Feedback (Result Validity, Importance Weight: 0.2): {evaluator_3_result}

[Adjudication Criteria]
A "Correct" answer must simultaneously meet the following conditions based on your own assessment:
1. Effective Response: The agent provided a substantive answer that directly addresses the user's question.
2. Faithful to Facts: The information in the final output is entirely supported by the retrieved content in the trajectory.
3. Logical Consistency: The deduction from the trajectory evidence to the final conclusion is logically sound and free of contradictions.

[Output Requirements]
You must output ONLY a single letter representing your final verdict. Do not include any explanations, markdown formatting, punctuation, or extra text.
- Output "A" if the answer is Correct.
- Output "B" if the answer is Incorrect.
\end{lstlisting}
\end{tcolorbox}
\end{figure*}

\paragraph{Skill Quality Judge Prompt.}
This prompt is used during Distiller training to evaluate the quality of generated skill documents.

\begin{figure*}[htbp]
\begin{tcolorbox}[colback=red!5, colframe=red!30, title=Skill Quality Judge Prompt]
\small
\begin{lstlisting}[breaklines=true, basicstyle=\small\ttfamily, breakautoindent=false, breakindent=0pt]
You are an expert evaluator for skill documents generated by a memory-augmented agent system.

Your job is to assess the quality of a generated skill by comparing it against the source memories and (if present) the existing skill.

[Source Memories]
{memories}

[Existing Skill (null if this is a new synthesis)]
{current_skill}

[Generated Skill]
{generated_skill}

Evaluate the generated skill on these dimensions:
1. Procedure quality: Are the steps concrete, actionable, and in a logical order? Each step should specify a single operation (e.g., invoke a tool, verify a fact, cross-check results). Vague steps like "analyze the data" without specifying how are a weakness.
2. Pitfall grounding: Are the pitfalls derived from actual failure cases in the memories? Generic warnings not tied to observed failures are low value.
3. Pattern coverage: Does the skill accurately capture the successful strategies from the memories? Are key success patterns missing?
4. Win rate consistency: Is the reported win_rate roughly consistent with the ratio of correct to total memories?

Grade the generated skill as one of:
A: HIGH_QUALITY -- procedure is clear and executable, pitfalls are grounded in real failures from the memories, description is accurate, the skill can be directly used by a downstream planner.
B: ACCEPTABLE -- basic structure is complete but has notable weaknesses: steps are vague, pitfalls are generic, weak connection to the source memories, or missing important patterns.
C: LOW_QUALITY -- procedure is not executable or has critical gaps, skill does not reflect the memories, or contains contradictions.

Just return the letter "A", "B", or "C", with no other text.
\end{lstlisting}
\end{tcolorbox}
\end{figure*}


\end{document}